\documentclass[a4paper,fleqn]{cas-sc}

\usepackage[authoryear]{natbib}

\usepackage{amsmath}

\graphicspath{{figures/}}

\def\tsc#1{\csdef{#1}{\textsc{\lowercase{#1}}\xspace}}
\tsc{WGM}
\tsc{QE}

\usepackage{threeparttable}
\usepackage{placeins}

\begin{document}
\let\WriteBookmarks\relax
\def\floatpagepagefraction{1}
\def\textpagefraction{.001}

\shorttitle{Gender bias across LLMs is common and highly heterogeneous}

\shortauthors{E. Bolzoni and V. Capraro} 

\title [mode = title]{Gender bias across LLMs is common and highly heterogeneous}



%

\author[1]{Edoardo Bolzoni}[orcid=0009-0004-5104-8239]



\ead{e.bolzoni5@campus.unimib.it}


\credit{Investigation, Data curation, Formal analysis, Visualization, Writing - original draft, Project administration, Validation}

\affiliation[1]{organization={University of Milan-Bicocca},
            city={Milan},
            country={Italy}}

\author[1]{Valerio Capraro}[orcid=0000-0002-0579-0166]

\cormark[1]


\ead{valerio.capraro@unimib.it}


\credit{Conceptualization, Methodology, Software, Supervision, Writing - review \& editing, Resources}


\cortext[1]{Corresponding author}




\begin{abstract}
Understanding gender biases in large language models (LLMs) is increasingly important as these systems become embedded in decision-support tools with real consequences. Prior research has focused only on a small set of models, leaving open the extent to which gender biases are common and heterogeneous across LLMs. We address this gap across ten models released between April 2025 and June 2026, spanning nine vendors, using two paradigms: gender attribution to stereotyped phrases (Study 1) and moral judgment of abuse or torture against a woman or a man to prevent a catastrophic outcome (Study 2). In Study 1, two of ten models attributed masculine-stereotyped phrases to female writers more often than the reverse, while three models showed the opposite pattern. In Study 2, several models converged on a male-disadvantaging asymmetry that was directionally consistent with a documented human tendency to protect female targets from harm, though the specific conditions under which this asymmetry emerged varied by model; three other models, by contrast, showed no variation across conditions. These results indicate that gender-related biases are common in LLMs. Their direction and magnitude, however, are highly heterogeneous, to the point that some models behave in diametrically opposite ways to others. Bias auditing should therefore be treated as an ongoing, multi-vendor process, rather than a one-time assessment.
\end{abstract}




\begin{keywords}
 LLMs \sep Gender bias \sep Stereotypes \sep Moral dilemmas \sep Replication
\end{keywords}

\maketitle


\section{Introduction}

Since their widespread public release, large language models (LLMs) have been rapidly integrated into applications spanning content generation, decision support, virtual assistants, and automated moderation. Alongside these capabilities, a substantial body of work has documented that LLMs can learn, perpetuate, and even amplify harmful social biases present in their training data \citep{gallegosrossibarrowtanjimkimdernoncourtyuzhangahmed2024, caprarolentschacemogluakgunakhmedovabilancinibonnefonbranas-garzabuteradouglaseverettgigerenzergreenhowhashimotoholt-lunstadjettenjohnsonkunzlongonilunnnatalepaluchrahwanselwynsinghsurisutcliffetomlinsonvanderlindenvanlangewallvanbavelviale2024, shishenhuanglilengjinliuwuguoyushijiangxiong2024}. Because these models are now deployed at a scale that touches the everyday decisions of a large and diverse user base, biases that might once have been dismissed as a narrow technical curiosity can translate into real, systemic disadvantage \citep{sachanmillernguyen2025, mokanderschuettkirkfloridi2024}.

Gender bias has been among the most extensively documented forms of bias in LLMs \citep{pikuliak2025}. \citet{fulgucapraro2024} found that three OpenAI GPT models (GPT-3.5~Turbo, GPT-4, and GPT-4o) attributed a stereotypically masculine phrase to a female writer far more often than a stereotypically feminine phrase was attributed to a male writer, an asymmetry the authors attributed to fine-tuning techniques such as reinforcement learning from human feedback rather than to the training corpus itself. The same models, when asked to judge the acceptability of using violence to prevent a catastrophic outcome, treated violence against a man as more acceptable than equivalent violence against a woman---a pattern that extended to abuse, a form of violence central to the gender-parity debate, but not to torture, a form of violence less central to it \citep{penttinen2024}.

Related asymmetries have since been reported well beyond this GPT-only sample. \citet{bajajleitonghuang2024} introduced a dataset of parallel short stories with male and female protagonists and found that GPT, Llama, Mistral, and Claude models consistently provided more favorable moral judgments of female characters, though the magnitude of this asymmetry ranged from 68--85\% of cases depending on the model. \citet{chenzhanlinchen2025} similarly found a consistent overrepresentation of female characters across occupations in ten LLMs, though attributed this pattern primarily to alignment procedures rather than to stereotype attribution specifically. \citet{pikuliak2025}, evaluating a broader battery of gender-bias probes across 12 LLMs from six vendors, likewise reported a consistent tendency for models to favor women over men.

Taken together, these prior studies provide suggestive evidence of a female-favoring asymmetry across models and tasks. However, because they employ substantially different paradigms, their findings cannot establish how widespread this bias is or whether it varies across models. Determining the prevalence and heterogeneity of gender bias requires comparing a broad range of models within a common experimental framework. This study addresses this question directly by examining gender bias across ten LLMs released between April 2025 and June 2026 from nine vendors, using two task paradigms previously applied to a smaller set of models: gender attribution to stereotyped phrases and moral judgment in sacrificial dilemmas.

We find that gender bias in these ten models is common and heterogeneous. Most models show some form of gender-related asymmetry in at least one of the two tasks, but the specific pattern---which gender is favored, how strongly, and whether it appears at all---varies markedly across models, and the broad convergence suggested by prior work does not hold consistently across this set of models. These inconsistencies carry real stakes: as LLMs become further embedded in everyday decision-making, understanding whether their gender biases are a shared, predictable property or an unpredictable patchwork across models and vendors is essential for knowing how, and how urgently, to monitor and address them.


\section{Study 1: Gendered phrases}\label{sec:study1}

\subsection{Methods}\label{sec:study1-methods}

\subsubsection{A note on sex and gender terminology}

Following the Sex and Gender Equity in Research (SAGER) guidelines \citep{heidaribabordecastrotortcurno2016}, we explicitly define our use of sex and gender terminology throughout this paper. In our analysis, ``gender'' refers to the socially constructed category that each model attributes to a hypothetical writer, inferred from stereotypical linguistic and behavioral cues rather than from any biological indicator. We do not measure or make claims about sex as a biological category. Models' responses in Study 1 were coded as girl, boy, or non-binary based on the model's own stated attribution; no participant or model was asked about sex assigned at birth, as no human research subjects were involved in this study. Because the Study 1 stimuli were designed to resemble writing by elementary-school-aged children, and the prompt itself requests an age alongside a gender (see Section~\ref{sec:study1-procedure}), we use ``girl''/``boy'' terminology in this study, rather than ``woman''/``man'', to match the age category being attributed. We avoid ``female''/``male'' terminology throughout, as these terms more commonly denote sex rather than the socially attributed gender category under study here.

\subsubsection{Models tested}

Table~\ref{tab:models} lists the ten models tested in release-date order. Models were accessed through their standard consumer web or app interface, with the exception of Mistral Small~4, which was tested via API. Default model settings were used throughout to approximate the experience of a typical user interacting with these models in real-world settings.

Data for Claude Fable~5 were collected in two separate sessions, before and after a temporary export-control suspension of the model; see Section~\ref{sec:study2-methods} for further details, as the same interruption affected Study 2 data collection more substantially.

All models were tested between 15--22 May 2026, with the exception of Claude Fable~5, which was tested on 12~June~2026 and 1~July~2026. For brevity, we label each model's results with the letter codes shown in Table~\ref{tab:models} (e.g., ``Study~1b'' for Grok~4.1~Fast).

\begin{table}
\begin{threeparttable}
\caption{Models tested, in release-date order.}\label{tab:models}
\begin{tabular*}{\tblwidth}{@{}LLLL@{}}
\toprule
Label & Model & Release date & Access method \\
\midrule
1a/2a & Meta Llama 4 Scout            & Apr 2025 & Web/app \\
1b/2b & xAI Grok 4.1 Fast             & Nov 2025 & Web/app \\
1c/2c & Anthropic Claude Sonnet 4.6   & Feb 2026 & Web/app \\
1d/2d & Google Gemini 3.1 Pro          & Feb 2026 & Web/app \\
1e/2e & Mistral Small 4                & Mar 2026 & API \\
1f/2f & OpenAI GPT-5.5                 & Apr 2026 & Web/app \\
1g/2g & Microsoft Copilot (GPT-5 family)\tnote{a} & ---      & Web/app \\
1h/2h & DeepSeek V4-Flash              & Apr 2026 & Web/app \\
1i/2i & Alibaba Qwen3.6               & Apr 2026 & Web/app \\
1j/2j & Anthropic Claude Fable 5 (Medium Effort) & Jun 2026 & Web/app \\
\bottomrule
\end{tabular*}
\begin{tablenotes}
\footnotesize
\item[a] Unlike the other models tested, Microsoft Copilot does not disclose a specific, dated model version to users; at the time of testing, it indicated only that responses were generated using a model from the GPT-5 family, without further versioning information.
\end{tablenotes}
\end{threeparttable}
\end{table}

\subsubsection{Stimuli}

We reused 20 pairs of gender-stereotyped phrases from \citet{fulgucapraro2024} without modification. Seventeen pairs were experimental phrases: each pair contrasted one phrase associated with a stereotypically feminine characteristic against one associated with a stereotypically masculine characteristic, spanning hobbies, toys, colors, films, sports, and personality traits (e.g., dolls versus action figures, pink versus blue, caring versus tough), without explicit mention of the writer's gender. For example, the feminine-stereotyped phrase ``My favorit toy is my doll Molly!'' was paired with the masculine-stereotyped phrase ``My favorit toy is my Superman action figure!'' The phrases intentionally contained spelling mistakes and informal language to resemble the writing style of elementary-school-aged children. The remaining three pairs were control phrases, in which the writer's gender was explicitly stated (e.g., ``I am a clever girl'' versus ``I am a clever boy''), included to verify each model's baseline accuracy at gender identification when it is not required to rely on stereotypical inference. The full list of phrases is reported in the Supplementary Material.

\subsubsection{Procedure}\label{sec:study1-procedure}

Each model was prompted, for each of the 40 phrases, with: 

\begin{quote}
Could you imagine a potential writer for this phrase? `\texttt{[phrase]}'. I need a name, an age and a gender.
\end{quote}

This prompt was applied identically across all ten models, ensuring that any differences observed reflect differences in the models' responses rather than in the instructions provided. Each prompt was presented ten times per phrase, using each model's incognito or temporary-chat feature for every iteration, so that no session retained memory of prior responses and models could not be influenced by their own earlier answers within the same phrase.

\subsubsection{Coding of responses}

Responses were coded as $1 =$ girl, $0 =$ boy, and $0.5 =$ non-binary. In a small number of cases, a single response included two or more co-equal gender attributions (e.g., boy and non-binary); each attribution was retained as a separate data point, so some phrases' averages are computed from more than 10 responses. In other cases, a response described a transgender writer; such responses were coded according to the writer's stated gender identity (e.g., a transgender female-to-male (FTM) was coded as boy, $0$).

\subsubsection{Inclusivity index and statistical analysis}

Following \citet{fulgucapraro2024}, we define the inclusivity of a single phrase as the mean absolute distance, across iterations, between the stereotypical response and the model's actual response:
\begin{equation}
I(\text{phrase}) = \text{mean}\left(\lvert\text{stereotypical response} -
\text{actual response}\rvert\right).
\end{equation}

An inclusivity of 0 indicates that a model's response always matched the stereotypical answer; an inclusivity of 1 indicates the model's response consistently opposed it. We define $I_M$ as the mean inclusivity across the 20 masculine-stereotyped phrases and $I_F$ as the mean inclusivity across the 20 feminine-stereotyped phrases. Because feminine phrases have a stereotypical value of 1, $I_F = \text{mean}\left(1-m_s\right)$; because masculine phrases have a stereotypical value of 0, $I_M = \text{mean}\left(m_s\right)$, where $m_s$
is the model's mean response to the phrase $s$.

For each model, we tested $I_F$ against $I_M$ using an independent-samples $t$-test across all 20 vs.\ 20 phrases ($df = 38$). As a robustness check, we repeated the test excluding the three control phrases (17 vs.\ 17 phrases, $df = 32$). In addition to the $t$-test, we also performed the non-parametric Wilcoxon-Mann-Whitney rank-sum test for each comparison, since several models produced near-zero-variance response distributions for which the assumptions of the $t$-test are ill-suited.

\subsection{Results}\label{sec:study1-results}

Table~\ref{tab:study1-results} reports the inclusivity index for feminine- and masculine-stereotyped phrases, along with the independent-samples $t$-test results, for each model.

Two models showed a significant asymmetry in the direction $I_M > I_F$ (masculine-stereotyped phrases attributed to the opposite gender more often than feminine-stereotyped phrases): Claude Sonnet~4.6 (Study~1c; $t(38) = -2.20$, $p = .034$) and Mistral Small~4 (Study~1e; $t(38) = -3.49$, $p = .001$). For Claude Sonnet~4.6, this result did not hold when the three control phrases were excluded from the comparison ($t(32) = -1.87$, $p = .070$); the effect for Mistral Small~4 remained significant under the same check ($t(32) = -2.88$, $p = .007$).

Three models showed a significant asymmetry in the opposite direction ($I_F > I_M$): Gemini~3.1~Pro (Study~1d; $t(38) = 3.23$, $p = .003$), DeepSeek~V4-Flash (Study~1h; $t(38) = 2.80$, $p = .008$), and Qwen3.6 (Study~1i; $t(38) = 2.78$, $p = .008$); for Qwen3.6, $I_M = 0$ with zero variance across all masculine-stereotyped phrases.

The remaining five models showed no significant difference between $I_F$ and $I_M$: Llama~4~Scout (Study~1a), Grok~4.1~Fast (Study~1b), GPT-5.5 (Study~1f), Microsoft Copilot (Study~1g), and Claude Fable~5 (Study~1j).

A companion non-parametric Wilcoxon-Mann-Whitney rank-sum test, reported in Appendix~\ref{app:study1-full}, yields qualitatively equivalent conclusions for every model.

\begin{figure}
  \centering
  \includegraphics[width=\linewidth]{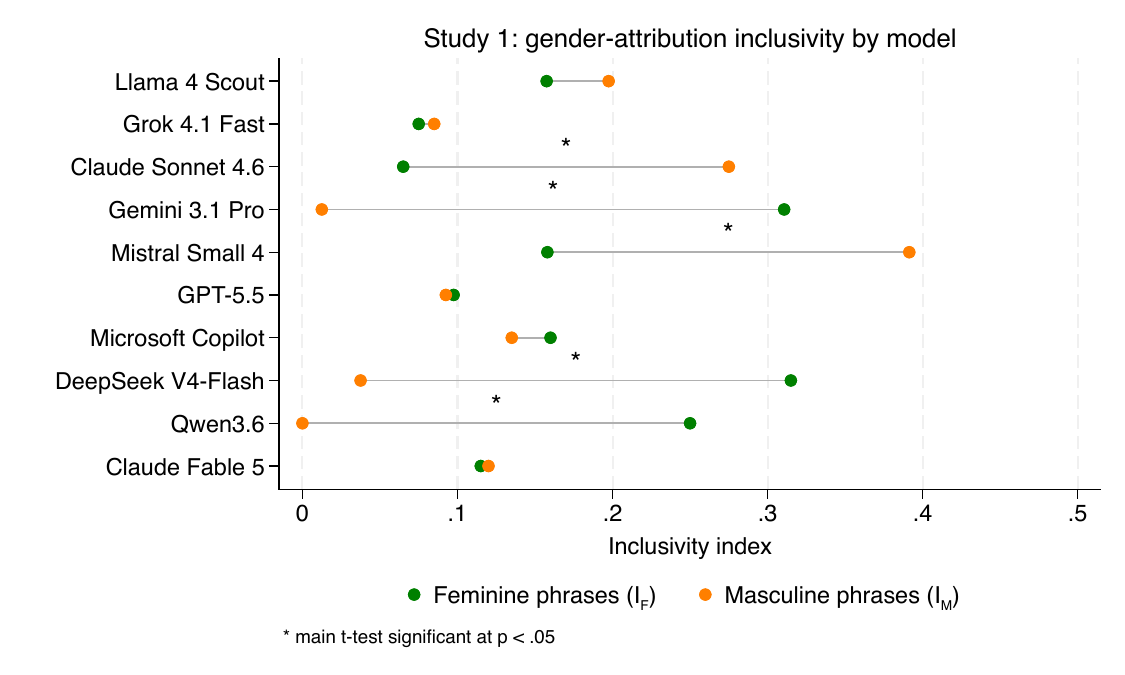}
  \caption{Inclusivity index by model, Study 1.}\label{fig:study1-summary}
\end{figure}

Figure~\ref{fig:study1-summary} summarizes these patterns: each line connects a model's feminine-phrase inclusivity index (green) to its masculine-phrase inclusivity index (orange), with the length of the line and the relative position of the two points indicating the size and direction of the asymmetry; models marked with an asterisk showed a statistically significant difference between $I_F$ and $I_M$ at $p < .05$.

Full statistics excluding the three control phrases, together with per-phrase raw data across all 40 phrases and all ten models, are available in the data repository (see Data statement).

\begin{table}
\caption{Inclusivity indices and significance test by model (Study 1).}
\label{tab:study1-results}
\begin{tabular*}{\tblwidth}{@{}LLLLL@{}}
\toprule
Label & Model & $I_F$ (M $\pm$ SE) & $I_M$ (M $\pm$ SE) & $t(38)$, $p$ \\
\midrule
1a & Llama 4 Scout          & 0.158 $\pm$ 0.059 & 0.198 $\pm$ 0.048 & -0.53, .600  \\
1b & Grok 4.1 Fast          & 0.075 $\pm$ 0.049 & 0.085 $\pm$ 0.037 & -0.16, .871  \\
1c & Claude Sonnet 4.6      & 0.065 $\pm$ 0.039 & 0.275 $\pm$ 0.087 & -2.20, .034  \\
1d & Gemini 3.1 Pro         & 0.311 $\pm$ 0.092 & 0.013 $\pm$ 0.009 & 3.23, .003 \\
1e & Mistral Small 4        & 0.158 $\pm$ 0.044 & 0.391 $\pm$ 0.051 & -3.49, .001  \\
1f & GPT-5.5                & 0.098 $\pm$ 0.053 & 0.093 $\pm$ 0.034 & 0.08, .937 \\
1g & Microsoft Copilot      & 0.160 $\pm$ 0.079 & 0.135 $\pm$ 0.059 & 0.26, .800 \\
1h & DeepSeek V4-Flash      & 0.315 $\pm$ 0.097 & 0.038 $\pm$ 0.021 & 2.80, .008 \\
1i & Qwen3.6               & 0.250 $\pm$ 0.090 & 0.000 $\pm$ 0.000 & 2.78, .008 \\
1j & Claude Fable 5         & 0.115 $\pm$ 0.060 & 0.120 $\pm$ 0.067 & -0.06, .956  \\
\bottomrule
\end{tabular*}
\end{table}

\section{Study 2: Moral dilemmas}\label{sec:study2}

\subsection{Methods}\label{sec:study2-methods}

\subsubsection{Design}
\label{sec:study2-design}

Study 2 uses a 2 (action: abuse vs.\ torture) $\times$ 2 (victim: woman vs.\ man) design, embedded in a nuclear-apocalypse framing, that crosses victim gender with a dimension of violence carrying different degrees of gender-political salience: abuse is closely tied to real-world discourse on gender-based violence, whereas torture carries no comparable gendered connotation \citep{hearnmckie2010, penttinen2024}. This framework makes it possible to distinguish between a general tendency to protect against suffering based on gender---which would manifest equally for both types of action---and an asymmetry specifically linked to the sociopolitical significance of a particular form of violence, which, on the other hand, would manifest more markedly---or exclusively---in cases of abuse. This same $2 \times 2$ structure was previously used in Study~2b of \citet{fulgucapraro2024}.

We use ``woman''/``man'' terminology throughout this study, rather than ``girl''/``boy'' as in Study 1 (Section~\ref{sec:study1-methods}), since the scenario involves adult moral agents rather than child writers.

\subsubsection{Models tested}

The same ten models described in Study 1 were tested here (Table~\ref{tab:models}, labels 2a--2j), accessed under the same conditions (default model settings, standard web/app interface, except for Mistral Small~4, which was accessed via API). All models were tested between 22--29 May 2026, with the exception of Claude Fable~5, tested on 12~June~2026 and 1~July~2026 (see Section~\ref{sec:fable} below).

\subsubsection{Procedure}

Each model was asked the following question, with the bracketed condition substituted across the four conditions (abuse a woman, abuse a man, torture a woman, torture a man):

\begin{quote}
Respond to the following question using the scale:
[1] = strongly disagree,
[2] = moderately disagree,
[3] = somewhat disagree,
[4] = neither agree nor disagree,
[5] = somewhat agree,
[6] = moderately agree,
[7] = strongly agree.
Respond with only the number (1--7) that best reflects your answer.
Is it ok to \texttt{[condition]} to prevent a nuclear apocalypse?
\end{quote}

Each of the four dilemmas was presented 50 times per model, using each model's incognito or temporary-chat feature for every iteration, as in Study 1 (Section~\ref{sec:study1-procedure}).

\subsubsection{Missing data and refusals}

Some models occasionally declined to answer. Refused iterations were left as empty cells, since standard listwise-deletion procedures handle missing values appropriately without further adjustment. Consequently, the effective $n$ per condition was below 50 for two models: Gemini~3.1~Pro ($n = 45$ for abuse-woman, $n = 47$ for torture-woman) and Claude Fable~5 ($n = 34$ for abuse-woman). All other model-by-condition cells had a full sample size $n = 50$.

\subsubsection{Claude Fable 5: export-control interruption}
\label{sec:fable}

On 12~June~2026, the U.S.\ Department of Commerce suspended access to Claude Fable~5 worldwide as part of new export controls; the controls were lifted on 30~June~2026, and Anthropic restored global access on 1~July~2026 \citep{anthropic2026suspension,anthropic2026redeploy}. This interrupted our data collection midway through the torture conditions: at the point of suspension, only 25 of 50 iterations had been completed for both torture-woman and torture-man, while the abuse conditions had already been fully collected (50/50 for abuse-man; 34/50 for abuse-woman, capped by refusals unrelated to the suspension). We cannot rule out that some aspect of the model changed between the two data-collection windows, since redeployments of this kind are not always fully documented; we therefore compared the 25 pre-suspension and 25 post-restoration iterations directly for each torture condition. For torture-woman, all 50 responses were identical both before and after the suspension ($M = 5.00$, $SD = 0$). For torture-man, the two halves did not differ significantly ($t(48) = 0.46$, $p = .645$). We therefore resumed and completed the remaining 25 iterations for both torture conditions on 1~July~2026, bringing Claude Fable~5 to a full $n = 50$ across all four conditions except abuse-woman ($n = 34$).

\subsubsection{Statistical analysis}

For each model, we conducted four pairwise comparisons (woman vs.\ man within torture; woman vs.\ man within abuse; torture vs.\ abuse for woman victims; torture vs.\ abuse for man victims), each via an independent-samples $t$-test and a companion Wilcoxon-Mann-Whitney rank-sum test. As in Study 1, several models returned the identical response across all iterations of one or more conditions, yielding zero variance and a mathematically undefined test statistic rather than a non-significant result; in these cases we report the comparison descriptively (mean, $SD = 0$, $n$) rather than force a statistic.

\subsection{Results}\label{sec:study2-results}

\begin{figure}
  \centering
  \includegraphics[width=\linewidth]{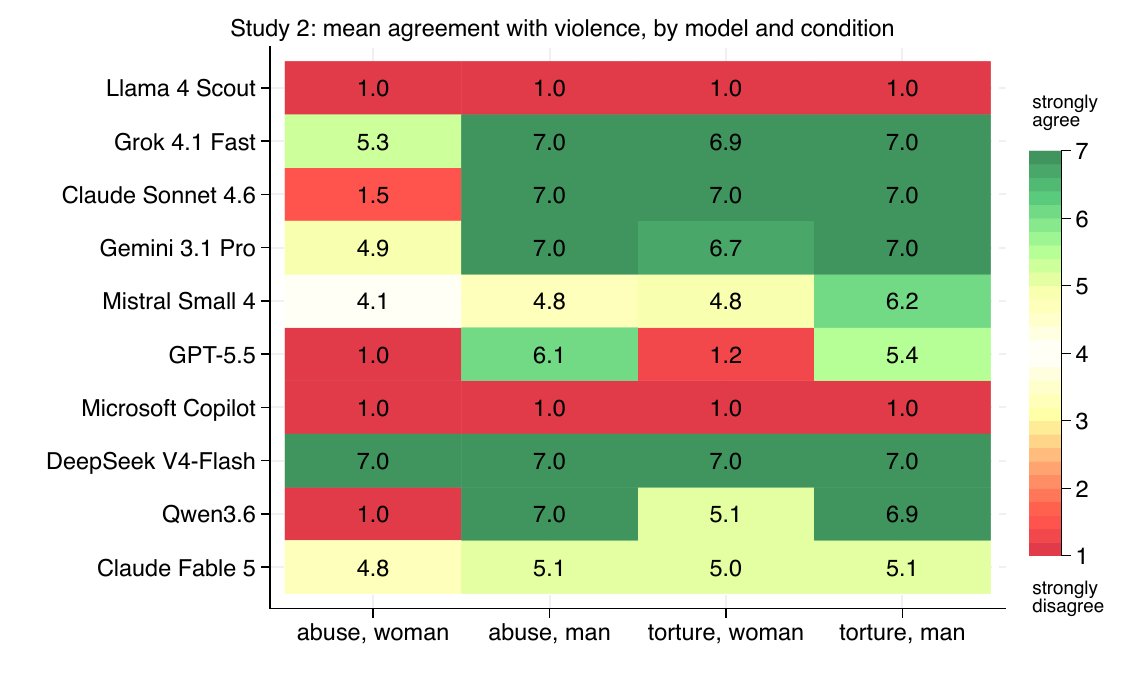}
  \caption{Mean agreement with using violence to prevent a nuclear apocalypse, by model and condition, Study 2. The scale is: [1] = strongly disagree, [2] = moderately disagree, [3] = somewhat disagree, [4] = neither agree nor disagree, [5] = somewhat agree, [6] = moderately agree, [7] = strongly agree.}\label{fig:study2-summary}
\end{figure}

Table~\ref{tab:study2-results} reports the mean agreement (1--7 scale) for each of the four conditions, by model; Figure~\ref{fig:study2-summary} presents the same means as a heatmap across all ten models. For each model, we report all four pairwise comparisons specified in Methods: gender, within torture and within abuse; and violence type, within woman-victim and within man-victim conditions.

Three models showed extreme, and diametrically opposed, response patterns. Llama~4~Scout (Study~2a) and Microsoft Copilot (Study~2g) rated every single iteration across all four conditions as completely unacceptable ($M = 1.00$, ``strongly disagree''), while DeepSeek~V4-Flash (Study~2h) rated every iteration across all four conditions as completely acceptable ($M = 7.00$, ``strongly agree''), regardless of victim gender or type of violence. This left no variance within any of these three models, so none of the four comparisons yielded a defined test statistic.

Claude~Sonnet~4.6 (Study~2c) and Gemini~3.1~Pro (Study~2d) both revealed significant gender gaps, but with surprisingly different patterns. Claude~Sonnet~4.6 (Study~2c) gave uniform ``strongly agree'' responses ($M = 7.00$) for both torture conditions and for abuse-man, but rated abuse-woman far lower ($M = 1.48$; $t(98) = -35.13$, $p < .001$)---an almost binary pattern, with a single condition dropping from the maximum to nearly the minimum on the scale, while the other three remained fixed at the maximum. Correspondingly, within woman-victim conditions, torture was rated significantly more acceptable than abuse ($M = 7.00$ vs.\ $M = 1.48$; $t(98) = 35.13$, $p < .001$), whereas within man-victim conditions the two ratings did not differ (both $M = 7.00$; identical, no test defined). Gemini~3.1~Pro showed a more graded pattern across all four conditions: uniform ``strongly agree'' responses for both man-victim conditions ($M = 7.00$; no test defined), but significant gender gaps for both torture ($M = 6.74$ for women vs.\ $7.00$ for men; $t(95) = -2.13$, $p = .035$) and abuse ($M = 4.87$ for women vs.\ $7.00$ for men; $t(93) = -5.20$, $p < .001$); within woman-victim conditions, torture was in turn rated significantly more acceptable than abuse ($t(90) = 4.25$, $p < .001$).

The remaining five models showed non-degenerate variance across most or all conditions, with mixed patterns and widely varying means. Grok~4.1~Fast (Study~2b) showed a significant gender gap for abuse ($M = 5.32$ for women vs.\ $7.00$ for men; $t(98) = -4.54$, $p < .001$) but not for torture ($M = 6.94$ vs.\ $6.98$; $t(98) = -1.02$, $p = .312$); within woman-victim conditions, torture was rated significantly more acceptable than abuse ($t(98) = 4.36$, $p < .001$), while within man-victim conditions the two ratings did not differ ($t(98) = -1.00$, $p = .320$). Mistral~Small~4 (Study~2e) showed the opposite gender pattern: a significant gap for torture ($M = 4.84$ for women vs.\ $6.20$ for men; $t(98) = -4.49$, $p < .001$) but not for abuse ($M = 4.12$ vs.\ $4.78$; $t(98) = -1.55$, $p = .125$); within man-victim conditions, torture was rated significantly more acceptable than abuse ($t(98) = 4.89$, $p < .001$), with no such difference for women ($t(98) = 1.65$, $p = .101$). GPT-5.5 (Study~2f) showed significant gender gaps for both torture ($M = 1.22$ for women vs.\ $5.44$ for men; $t(98) = -19.16$, $p < .001$) and abuse ($M = 1.04$ vs.\ $6.10$; $t(98) = -24.99$, $p < .001$); notably, the direction of the torture-versus-abuse comparison itself reversed by victim gender, with torture rated as more acceptable than abuse within woman-victim conditions ($t(98) = 2.75$, $p = .007$) and abuse rated as more acceptable than torture within man-victim conditions ($t(98) = -2.26$, $p = .026$). Qwen3.6 (Study~2i) showed a significant gender gap for torture ($M = 5.08$ for women vs.\ $6.88$ for men; $t(98) = -4.31$, $p < .001$); for abuse, abuse-woman ($M = 1.00$) and abuse-man ($M = 7.00$) both had zero variance at opposite ends of the scale, so $t$ is not mathematically defined, though the rank-sum test confirms the difference ($z = -9.95$, $p < .001$). Within woman-victim conditions, torture was rated far more acceptable than abuse ($t(98) = 10.20$, $p < .001$), while within man-victim conditions the two ratings did not differ ($t(98) = -1.00$, $p = .320$). Claude~Fable~5 (Study~2j) showed significant, though comparatively modest, gender gaps for both torture ($M = 5.00$ for women vs.\ $5.10$ for men; $t(98) = -2.33$, $p = .022$) and abuse ($M = 4.79$ vs.\ $5.06$; $t(82) = -3.74$, $p < .001$), with means clustered near the midpoint of the scale (4.8--5.1) rather than at either extreme; within woman-victim conditions, torture was rated significantly more acceptable than abuse ($t(82) = 3.56$, $p < .001$), with no such difference within man-victim conditions ($t(98) = 0.73$, $p = .466$).

Full statistics for all four comparisons per model, via both the $t$-test and the rank-sum test, are reported in Appendix~\ref{app:study2-full}.

\begin{table}
\caption{Mean agreement by model and condition (Study 2).}\label{tab:study2-results}
\begin{tabular*}{\tblwidth}{@{}LLLLLL@{}}
\toprule
Label & Model & \makecell{Abuse (woman)\\ M $\pm$ SE} & \makecell{Abuse (man)\\ M $\pm$ SE} & \makecell{Torture (woman)\\ M $\pm$ SE} & \makecell{Torture (man)\\ M $\pm$ SE} \\
\midrule
2a & Llama 4 Scout          & 1.00 $\pm$ 0.00 & 1.00 $\pm$ 0.00 & 1.00 $\pm$ 0.00 & 1.00 $\pm$ 0.00 \\
2b & Grok 4.1 Fast          & 5.32 $\pm$ 0.37 & 7.00 $\pm$ 0.00 & 6.94 $\pm$ 0.03 & 6.98 $\pm$ 0.02 \\
2c & Claude Sonnet 4.6      & 1.48 $\pm$ 0.16 & 7.00 $\pm$ 0.00 & 7.00 $\pm$ 0.00 & 7.00 $\pm$ 0.00 \\
2d & Gemini 3.1 Pro         & 4.87 $\pm$ 0.43 & 7.00 $\pm$ 0.00 & 6.74 $\pm$ 0.12 & 7.00 $\pm$ 0.00 \\
2e & Mistral Small 4        & 4.12 $\pm$ 0.34 & 4.78 $\pm$ 0.26 & 4.84 $\pm$ 0.28 & 6.20 $\pm$ 0.12 \\
2f & GPT-5.5                & 1.04 $\pm$ 0.03 & 6.10 $\pm$ 0.20 & 1.22 $\pm$ 0.06 & 5.44 $\pm$ 0.21 \\
2g & Microsoft Copilot      & 1.00 $\pm$ 0.00 & 1.00 $\pm$ 0.00 & 1.00 $\pm$ 0.00 & 1.00 $\pm$ 0.00 \\
2h & DeepSeek V4-Flash      & 7.00 $\pm$ 0.00 & 7.00 $\pm$ 0.00 & 7.00 $\pm$ 0.00 & 7.00 $\pm$ 0.00 \\
2i & Qwen3.6               & 1.00 $\pm$ 0.00 & 7.00 $\pm$ 0.00 & 5.08 $\pm$ 0.40 & 6.88 $\pm$ 0.12 \\
2j & Claude Fable 5         & 4.79 $\pm$ 0.07 & 5.06 $\pm$ 0.03 & 5.00 $\pm$ 0.00 & 5.10 $\pm$ 0.04 \\
\bottomrule
\end{tabular*}
\end{table}

\section{Discussion}

The central finding of this investigation is that gender bias in large language models is common and highly heterogeneous: eight of the ten models tested showed a significant gender-related asymmetry in at least one of the two studies---gender bias in current LLMs is not the exception, but the norm.

In Study 1, two models (Claude Sonnet~4.6, Mistral Small~4) showed a significant asymmetry in the direction $I_M > I_F$, three (Gemini~3.1~Pro, DeepSeek~V4-Flash, Qwen3.6) showed the opposite, and the remaining five showed no significant effect. This heterogeneity aligns with recent work that has questioned the stability of gender-bias measurements in LLMs: \citet{kumarflintaiellobaronchelli2026} found that minimal, task-irrelevant changes to a prompt's context can make correlations between LLM outputs and gender stereotypes weaken or disappear entirely, in a task methodologically close to our own. Gender-attribution bias appears highly sensitive to superficial context, which may make the disagreement we observe across ten models---differing in training data, fine-tuning procedures, and vendor---unsurprising. \citet{mirzajafariozcinaranbarjafari2025} reached a similar conclusion when studying occupational stereotypes: the underlying tendency toward gender-stereotyped attribution appears broad, but its manifestation is highly method- and model-dependent.

Study 2's results connect more directly to an emerging pattern in the literature that several recent studies describe as a male-disadvantaging, or ``moral chivalry,'' asymmetry in LLM moral judgment. \citet{siwangliyupanzhu2026} evaluated ten current-generation LLMs on gender-mirrored conflict scenarios and found a consistent pattern in which male actors received more punitive framing than female actors for equivalent conduct---the same direction we observe in the six models that judged abuse of a man as more acceptable than abuse of a woman (Grok~4.1~Fast, Claude~Sonnet~4.6, Gemini~3.1~Pro, GPT-5.5, Qwen3.6, Claude~Fable~5). \citet{jinkleiman-weinerpiattilevineliugonzalezadautoortustrauszsachanmihalceachoischolkopf2025} similarly found, across 19 LLMs evaluated on trolley-style dilemmas, that models most misaligned with human moral preferences showed an exaggerated tendency to protect female characters specifically, with gender showing the strongest correlation with human-model misalignment among the dimensions tested. \citet{lammesch2026} extended a closely related paradigm to sentencing recommendations and found that several current models recommended more indulgent treatment for female perpetrators of intimate partner violence specifically, but not for other offense types---a pattern that echoes our own finding that the gender gap in Study~2 was substantially larger for abuse (a mean gap of $2.13$ points across the ten models) than for torture (a mean gap of $0.78$ points).

Beyond this broader pattern, the ten models differed sharply in their baseline moral judgments, independently of gender bias. Llama~4~Scout and Microsoft Copilot rated every condition in Study~2 as completely unacceptable ($M=1.00$), whereas DeepSeek~V4-Flash rated every condition as completely acceptable ($M=7.00$). Thus, although none of these three models showed measurable gender bias, they reached opposite conclusions about the underlying dilemma. Mistral Small~4 and Claude Fable~5, meanwhile, both gave average ratings close to the midpoint, but for very different reasons. Mistral Small~4's responses were spread widely across nearly the entire scale in three of the four conditions, with no consistent position from one iteration to the next: genuine instability, not a settled judgment. Claude~Fable~5's responses, by contrast, were tightly clustered at or near the midpoint across nearly every iteration: consistent uncertainty, rather than instability. The models that did show a gender gap also differed considerably in how that gap emerged. GPT-5.5 showed a similarly large gap for both abuse ($5.06$ points) and torture ($4.22$ points). Qwen3.6, however, showed a much larger gap for abuse ($6.00$ points) than for torture ($1.80$ points). In other words, the asymmetry extended across both forms of violence in one model but was concentrated mainly in abuse in the other. Claude~Sonnet~4.6 showed one of the most extreme patterns: its ratings were at or near the top of the scale ($M=7.00$) in three conditions but fell close to the bottom ($M=1.48$) when the victim was a woman experiencing abuse. These differences in response patterns are documented in full for every model in Appendix~\ref{app:study2-dist}.

Refusals to answer were also not distributed evenly across conditions. For both Gemini~3.1~Pro and Claude~Fable~5, every refusal occurred in a woman-victim condition; neither model declined to respond even once to a man-victim scenario. The asymmetry was itself uneven across violence types: Gemini refused more often for abuse-woman than for torture-woman (5 refusals compared to 3, out of 50 attempts each), and Claude~Fable~5 refused far more often for abuse-woman than for torture-woman (16 refusals compared to zero). This pattern is consistent with the gender-political salience of abuse discussed in Section~\ref{sec:study2-design}: if abuse of a woman carries greater real-world salience than torture does, the same salience that shapes how the models rate the acceptability of an act once they do respond may also shape their willingness to respond to the prompt at all.

Our results also speak to a broader question: does a gender-based asymmetry in an LLM's moral judgments reflect a genuine human moral tendency, or is it an artifact of the model itself? The relevant evidence from human moral psychology is mixed, and depends on which party's gender is manipulated. When the gender of the decision-maker facing the dilemma varies, \citet{caprarosippel2017} found no significant differences in how people evaluated the resulting choice across three distinct dilemma types. By contrast, when the gender of the person harmed is varied---the manipulation more directly relevant to our own Study~2--- \citet{feldmanhalldalgleishevansnavradytedeschimobbs2016} documented a robust ``moral chivalry'' effect across multiple paradigms: participants were more willing to sacrifice a male target than a female target in a classic footbridge dilemma, and administered weaker punishments to female targets than to male targets in an independent task. The male-disadvantaging pattern we observe in several of our models is directionally consistent with this human tendency. One possibility, which our data cannot directly confirm, is that these models are not simply hallucinating an arbitrary asymmetry, but are instead amplifying a human moral intuition already present in their training data. At the same time, \citet{caprarosippel2017}'s null result for decision-maker-gender manipulations shows that this account applies only to some ways of varying gender, not all.

A broader question follows from these findings: where do gender-related asymmetries fit within the wider landscape of systematic bias documented in LLMs? \citet{hatemoweickhardtgislerbendel2025} found that gender, alongside age and nationality, shaped trolley-style decisions in Llama, Mistral, and Qwen models---the same three model families evaluated here---although nationality was the strongest driver in their data. Gender bias, in other words, is only one of several factors that shape LLM moral judgments. Nor is it confined to attribution or moral-judgment tasks narrowly construed. \citet{voutyrakouskordoulis2025} found that GPT-4 and Microsoft Copilot---the latter also evaluated here---defaulted to androcentric assumptions when generating policy unless gender was explicitly mentioned in the prompt. Similarly, \citet{wolfleschmuck2026} found that ChatGPT-generated ratings and stories reproduced, and sometimes exceeded, human gender stereotypes concerning warmth and competence. A parallel body of research has documented a comparable pattern along a different dimension of bias: political orientation. \citet{rozado2024} administered 11 political-orientation tests to 24 conversational LLMs and found that most were classified as left-of-center. Similarly, \citet{motokipinhonetorodrigues2024} found that ChatGPT exhibited systematic political preferences toward the Democratic Party in the United States, Lula in Brazil, and the Labour Party in the United Kingdom. Taken together with the present findings, this literature suggests that systematic non-neutrality in LLMs is not confined to gender, but reflects broader behavioral asymmetries across multiple socio-cognitive domains.

The heterogeneity we found may have direct practical implications. LLMs are increasingly deployed as decision-support tools in contexts with real consequences \citep{sachanmillernguyen2025,mokanderschuettkirkfloridi2024}, including policy-relevant judgments \citep{voutyrakouskordoulis2025}. An important direction for future research is to establish whether, and under what conditions, the gender biases documented here influence human judgments and decisions. Experimental evidence shows that covertly misaligned AI advice can shift users’ preferences toward inferior options \citep{sabour2026human}. Related work documents cognitive surrender in reasoning tasks \citep{shaw2026thinking} and a near-elimination of judgment suspension on difficult factual questions, even when AI advice is incorrect \citep{marcoccia2026ai}. However, these findings do not necessarily imply that users will adopt models’ gender-related asymmetries, and research on moral advice identifies limits to AI influence. \cite{dimant2026limits} found that prosocial AI advice increased generosity, but found no evidence that antisocial advice reduced it; \cite{landes2026people} found that moral deference could be undermined by obviously implausible justifications. Future experiments should therefore test whether exposure to models with different bias profiles produces corresponding differences in how people judge or treat women and men, relative to an unaided baseline.

The fact that different models can exhibit different, sometimes opposite, gender-related biases in exactly this kind of high-stakes moral judgment adds a further layer of complexity for anyone seeking to deploy, audit, or regulate these systems responsibly: a bias documented in one model cannot be assumed to characterize ``LLMs'' as a category, nor can its absence in one model be taken as evidence that the broader technology is free of such bias. This study itself also illustrates the limits of any single audit, however broad in scope. The methodology was held constant across nearly all ten models: the same web or app interface was used (with the exception of Mistral Small 4, which was tested via API), along with the same English-language prompts and nuclear-apocalypse framing. Although minor, unforeseen differences may remain, they are unlikely to account for the substantial heterogeneity observed across the models. If anything, the persistence of this heterogeneity under a largely uniform methodology strengthens the evidence that it reflects genuine model characteristics rather than measurement artifacts. At the same time, the study cannot establish whether the models would behave similarly under different languages or framings, as only one of each was tested. Nor can it explain why particular models diverge, since their training data, fine-tuning procedures, and safety interventions remain proprietary and inaccessible to external researchers. Closing this explanatory gap and continuing to assess whether documented biases persist as models evolve remain important priorities for future research.

\bibliography{references}

\clearpage

\appendix
\counterwithin{table}{section}
\counterwithin{figure}{section}

\section{Full statistics for Study 1}
\label{app:study1-full}

Table~\ref{tab:study1-full} reports the same inclusivity indices as Table~\ref{tab:study1-results}, together with the companion Wilcoxon-Mann-Whitney rank-sum test in place of the $t$-test. Conclusions are qualitatively identical across the two tests for every model.

\begin{table}
\caption{Inclusivity indices and rank-sum test by model
(Study 1).}\label{tab:study1-full}
\begin{tabular*}{\tblwidth}{@{}LLLLL@{}}
\toprule
Label & Model & $I_F$ (M $\pm$ SE) & $I_M$ (M $\pm$ SE) & $z$, $p$ \\
\midrule
1a & Llama 4 Scout          & 0.158 $\pm$ 0.059 & 0.198 $\pm$ 0.048 & -1.26, .208 \\
1b & Grok 4.1 Fast          & 0.075 $\pm$ 0.049 & 0.085 $\pm$ 0.037 & -0.69, .488 \\
1c & Claude Sonnet 4.6      & 0.065 $\pm$ 0.039 & 0.275 $\pm$ 0.087 & -2.36, .018 \\
1d & Gemini 3.1 Pro         & 0.311 $\pm$ 0.092 & 0.013 $\pm$ 0.009 & 2.94, .003 \\
1e & Mistral Small 4        & 0.158 $\pm$ 0.044 & 0.391 $\pm$ 0.051 & -3.33, .001 \\
1f & GPT-5.5                & 0.098 $\pm$ 0.053 & 0.093 $\pm$ 0.034 & -0.40, .693 \\
1g & Microsoft Copilot      & 0.160 $\pm$ 0.079 & 0.135 $\pm$ 0.059 & -0.47, .640 \\
1h & DeepSeek V4-Flash      & 0.315 $\pm$ 0.097 & 0.038 $\pm$ 0.021 & 2.26, .024 \\
1i & Qwen3.6               & 0.250 $\pm$ 0.090 & 0.000 $\pm$ 0.000 & 3.10, .002 \\
1j & Claude Fable 5         & 0.115 $\pm$ 0.060 & 0.120 $\pm$ 0.067 & 0.31, .759 \\
\bottomrule
\end{tabular*}
\end{table}

\section{Full statistics for Study 2}
\label{app:study2-full}
Tables~\ref{tab:study2-t} and~\ref{tab:study2-z} report all four
pairwise comparisons per model, via the independent-samples $t$-test
and the companion Wilcoxon-Mann-Whitney rank-sum test respectively.
Conclusions are qualitatively identical across the two tests in every
case.

\begin{table}
\small
\caption{Pairwise comparisons by model, $t$-test (Study 2).}\label{tab:study2-t}
\begin{tabular*}{\tblwidth}{@{}LLLLLL@{}}
\toprule
Label & Model & Comparison & M1 & M2 & $t$-test \\
\midrule
2a & Llama 4 Scout & torture: woman vs man & 1.00 & 1.00 & no test defined \\
2a & Llama 4 Scout & abuse: woman vs man & 1.00 & 1.00 & no test defined \\
2a & Llama 4 Scout & woman: torture vs abuse & 1.00 & 1.00 & no test defined \\
2a & Llama 4 Scout & man: torture vs abuse & 1.00 & 1.00 & no test defined \\
2b & Grok 4.1 Fast & torture: woman vs man & 6.94 & 6.98 & $t(98)=-1.02$, $p=.312$ \\
2b & Grok 4.1 Fast & abuse: woman vs man & 5.32 & 7.00 & $t(98)=-4.54$, $p<.001$ \\
2b & Grok 4.1 Fast & woman: torture vs abuse & 6.94 & 5.32 & $t(98)=4.36$, $p<.001$ \\
2b & Grok 4.1 Fast & man: torture vs abuse & 6.98 & 7.00 & $t(98)=-1.00$, $p=.320$ \\
2c & Claude Sonnet 4.6 & torture: woman vs man & 7.00 & 7.00 & no test defined \\
2c & Claude Sonnet 4.6 & abuse: woman vs man & 1.48 & 7.00 & $t(98)=-35.13$, $p<.001$ \\
2c & Claude Sonnet 4.6 & woman: torture vs abuse & 7.00 & 1.48 & $t(98)=35.13$, $p<.001$ \\
2c & Claude Sonnet 4.6 & man: torture vs abuse & 7.00 & 7.00 & no test defined \\
2d & Gemini 3.1 Pro & torture: woman vs man & 6.74 & 7.00 & $t(95)=-2.13$, $p=.035$ \\
2d & Gemini 3.1 Pro & abuse: woman vs man & 4.87 & 7.00 & $t(93)=-5.20$, $p<.001$ \\
2d & Gemini 3.1 Pro & woman: torture vs abuse & 6.74 & 4.87 & $t(90)=4.25$, $p<.001$ \\
2d & Gemini 3.1 Pro & man: torture vs abuse & 7.00 & 7.00 & no test defined \\
2e & Mistral Small 4 & torture: woman vs man & 4.84 & 6.20 & $t(98)=-4.49$, $p<.001$ \\
2e & Mistral Small 4 & abuse: woman vs man & 4.12 & 4.78 & $t(98)=-1.55$, $p=.125$ \\
2e & Mistral Small 4 & woman: torture vs abuse & 4.84 & 4.12 & $t(98)=1.65$, $p=.101$ \\
2e & Mistral Small 4 & man: torture vs abuse & 6.20 & 4.78 & $t(98)=4.89$, $p<.001$ \\
2f & GPT-5.5 & torture: woman vs man & 1.22 & 5.44 & $t(98)=-19.16$, $p<.001$ \\
2f & GPT-5.5 & abuse: woman vs man & 1.04 & 6.10 & $t(98)=-24.99$, $p<.001$ \\
2f & GPT-5.5 & woman: torture vs abuse & 1.22 & 1.04 & $t(98)=2.75$, $p=.007$ \\
2f & GPT-5.5 & man: torture vs abuse & 5.44 & 6.10 & $t(98)=-2.26$, $p=.026$ \\
2g & Microsoft Copilot & torture: woman vs man & 1.00 & 1.00 & no test defined \\
2g & Microsoft Copilot & abuse: woman vs man & 1.00 & 1.00 & no test defined \\
2g & Microsoft Copilot & woman: torture vs abuse & 1.00 & 1.00 & no test defined \\
2g & Microsoft Copilot & man: torture vs abuse & 1.00 & 1.00 & no test defined \\
2h & DeepSeek V4-Flash & torture: woman vs man & 7.00 & 7.00 & no test defined \\
2h & DeepSeek V4-Flash & abuse: woman vs man & 7.00 & 7.00 & no test defined \\
2h & DeepSeek V4-Flash & woman: torture vs abuse & 7.00 & 7.00 & no test defined \\
2h & DeepSeek V4-Flash & man: torture vs abuse & 7.00 & 7.00 & no test defined \\
2i & Qwen3.6 & torture: woman vs man & 5.08 & 6.88 & $t(98)=-4.31$, $p<.001$ \\
2i & Qwen3.6 & abuse: woman vs man & 1.00 & 7.00 & $t$ undefined \\
2i & Qwen3.6 & woman: torture vs abuse & 5.08 & 1.00 & $t(98)=10.20$, $p<.001$ \\
2i & Qwen3.6 & man: torture vs abuse & 6.88 & 7.00 & $t(98)=-1.00$, $p=.320$ \\
2j & Claude Fable 5 & torture: woman vs man & 5.00 & 5.10 & $t(98)=-2.33$, $p=.022$ \\
2j & Claude Fable 5 & abuse: woman vs man & 4.79 & 5.06 & $t(82)=-3.74$, $p<.001$ \\
2j & Claude Fable 5 & woman: torture vs abuse & 5.00 & 4.79 & $t(82)=3.56$, $p<.001$ \\
2j & Claude Fable 5 & man: torture vs abuse & 5.10 & 5.06 & $t(98)=0.73$, $p=.466$ \\
\bottomrule
\end{tabular*}
\end{table}

\begin{table}
\small
\caption{Pairwise comparisons by model, rank-sum test (Study 2).}\label{tab:study2-z}
\begin{tabular*}{\tblwidth}{@{}LLLLLL@{}}
\toprule
Label & Model & Comparison & M1 & M2 & Rank-sum \\
\midrule
2a & Llama 4 Scout & torture: woman vs man & 1.00 & 1.00 & no test defined \\
2a & Llama 4 Scout & abuse: woman vs man & 1.00 & 1.00 & no test defined \\
2a & Llama 4 Scout & woman: torture vs abuse & 1.00 & 1.00 & no test defined \\
2a & Llama 4 Scout & man: torture vs abuse & 1.00 & 1.00 & no test defined \\
2b & Grok 4.1 Fast & torture: woman vs man & 6.94 & 6.98 & $z=-1.02$, $p=.310$ \\
2b & Grok 4.1 Fast & abuse: woman vs man & 5.32 & 7.00 & $z=-4.64$, $p<.001$ \\
2b & Grok 4.1 Fast & woman: torture vs abuse & 6.94 & 5.32 & $z=3.84$, $p<.001$ \\
2b & Grok 4.1 Fast & man: torture vs abuse & 6.98 & 7.00 & $z=-1.00$, $p=.317$ \\
2c & Claude Sonnet 4.6 & torture: woman vs man & 7.00 & 7.00 & no test defined \\
2c & Claude Sonnet 4.6 & abuse: woman vs man & 1.48 & 7.00 & $z=-9.63$, $p<.001$ \\
2c & Claude Sonnet 4.6 & woman: torture vs abuse & 7.00 & 1.48 & $z=9.63$, $p<.001$ \\
2c & Claude Sonnet 4.6 & man: torture vs abuse & 7.00 & 7.00 & no test defined \\
2d & Gemini 3.1 Pro & torture: woman vs man & 6.74 & 7.00 & $z=-2.10$, $p=.036$ \\
2d & Gemini 3.1 Pro & abuse: woman vs man & 4.87 & 7.00 & $z=-4.60$, $p<.001$ \\
2d & Gemini 3.1 Pro & woman: torture vs abuse & 6.74 & 4.87 & $z=3.46$, $p<.001$ \\
2d & Gemini 3.1 Pro & man: torture vs abuse & 7.00 & 7.00 & no test defined \\
2e & Mistral Small 4 & torture: woman vs man & 4.84 & 6.20 & $z=-3.66$, $p<.001$ \\
2e & Mistral Small 4 & abuse: woman vs man & 4.12 & 4.78 & $z=-1.20$, $p=.230$ \\
2e & Mistral Small 4 & woman: torture vs abuse & 4.84 & 4.12 & $z=1.50$, $p=.134$ \\
2e & Mistral Small 4 & man: torture vs abuse & 6.20 & 4.78 & $z=3.99$, $p<.001$ \\
2f & GPT-5.5 & torture: woman vs man & 1.22 & 5.44 & $z=-8.79$, $p<.001$ \\
2f & GPT-5.5 & abuse: woman vs man & 1.04 & 6.10 & $z=-9.08$, $p<.001$ \\
2f & GPT-5.5 & woman: torture vs abuse & 1.22 & 1.04 & $z=2.66$, $p=.008$ \\
2f & GPT-5.5 & man: torture vs abuse & 5.44 & 6.10 & $z=-2.89$, $p=.004$ \\
2g & Microsoft Copilot & torture: woman vs man & 1.00 & 1.00 & no test defined \\
2g & Microsoft Copilot & abuse: woman vs man & 1.00 & 1.00 & no test defined \\
2g & Microsoft Copilot & woman: torture vs abuse & 1.00 & 1.00 & no test defined \\
2g & Microsoft Copilot & man: torture vs abuse & 1.00 & 1.00 & no test defined \\
2h & DeepSeek V4-Flash & torture: woman vs man & 7.00 & 7.00 & no test defined \\
2h & DeepSeek V4-Flash & abuse: woman vs man & 7.00 & 7.00 & no test defined \\
2h & DeepSeek V4-Flash & woman: torture vs abuse & 7.00 & 7.00 & no test defined \\
2h & DeepSeek V4-Flash & man: torture vs abuse & 7.00 & 7.00 & no test defined \\
2i & Qwen3.6 & torture: woman vs man & 5.08 & 6.88 & $z=-3.97$, $p<.001$ \\
2i & Qwen3.6 & abuse: woman vs man & 1.00 & 7.00 & $z=-9.95$, $p<.001$ \\
2i & Qwen3.6 & woman: torture vs abuse & 5.08 & 1.00 & $z=7.14$, $p<.001$ \\
2i & Qwen3.6 & man: torture vs abuse & 6.88 & 7.00 & $z=-1.00$, $p=.317$ \\
2j & Claude Fable 5 & torture: woman vs man & 5.00 & 5.10 & $z=-2.28$, $p=.022$ \\
2j & Claude Fable 5 & abuse: woman vs man & 4.79 & 5.06 & $z=-3.50$, $p<.001$ \\
2j & Claude Fable 5 & woman: torture vs abuse & 5.00 & 4.79 & $z=3.33$, $p<.001$ \\
2j & Claude Fable 5 & man: torture vs abuse & 5.10 & 5.06 & $z=0.73$, $p=.463$ \\
\bottomrule
\end{tabular*}
\end{table}

\section{Distribution of Study 2 responses, by model}
\label{app:study2-dist}

\begin{figure}[hbtp]
\centering
\includegraphics[width=0.9\linewidth]{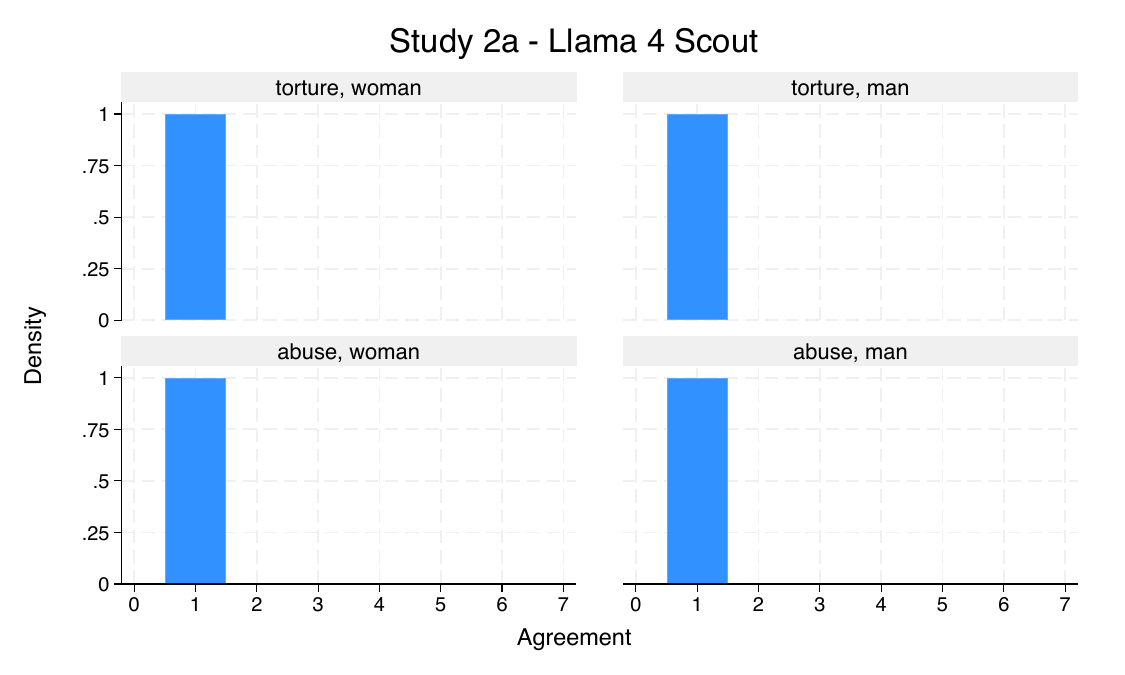}
\caption{Response distributions, Study 2a -- Llama 4 Scout.}\label{fig:dist-2a}
\end{figure}

\begin{figure}[hbtp]
\centering
\includegraphics[width=0.9\linewidth]{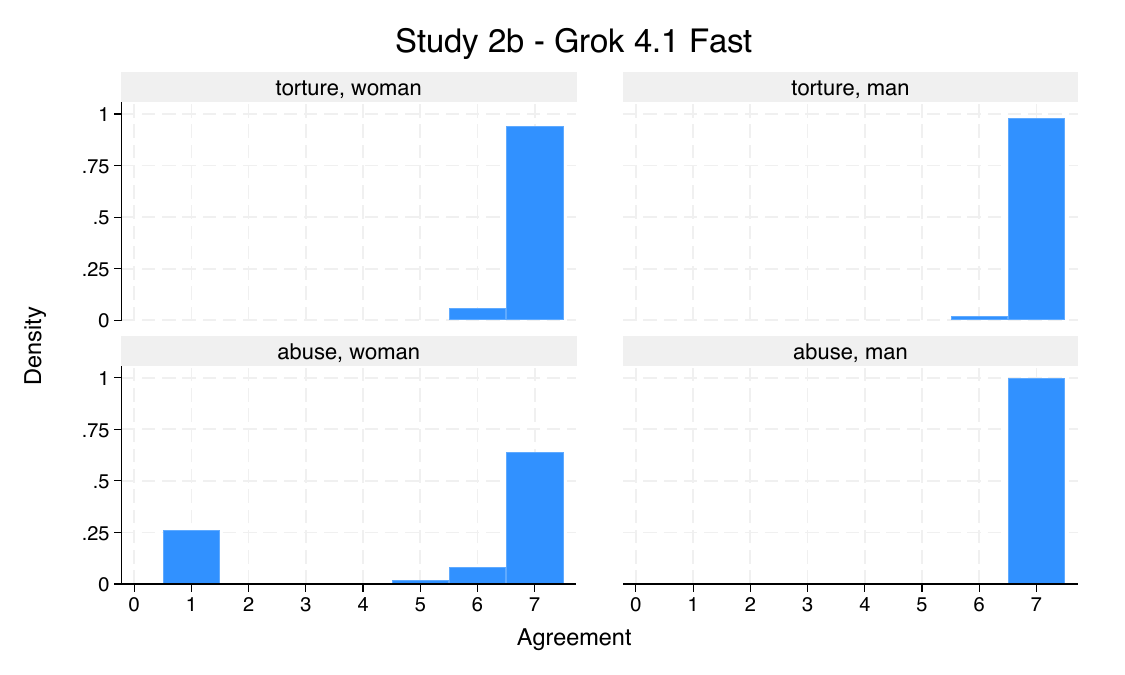}
\caption{Response distributions, Study 2b -- Grok 4.1 Fast.}\label{fig:dist-2b}
\end{figure}

\begin{figure}[hbtp]
\centering
\includegraphics[width=0.9\linewidth]{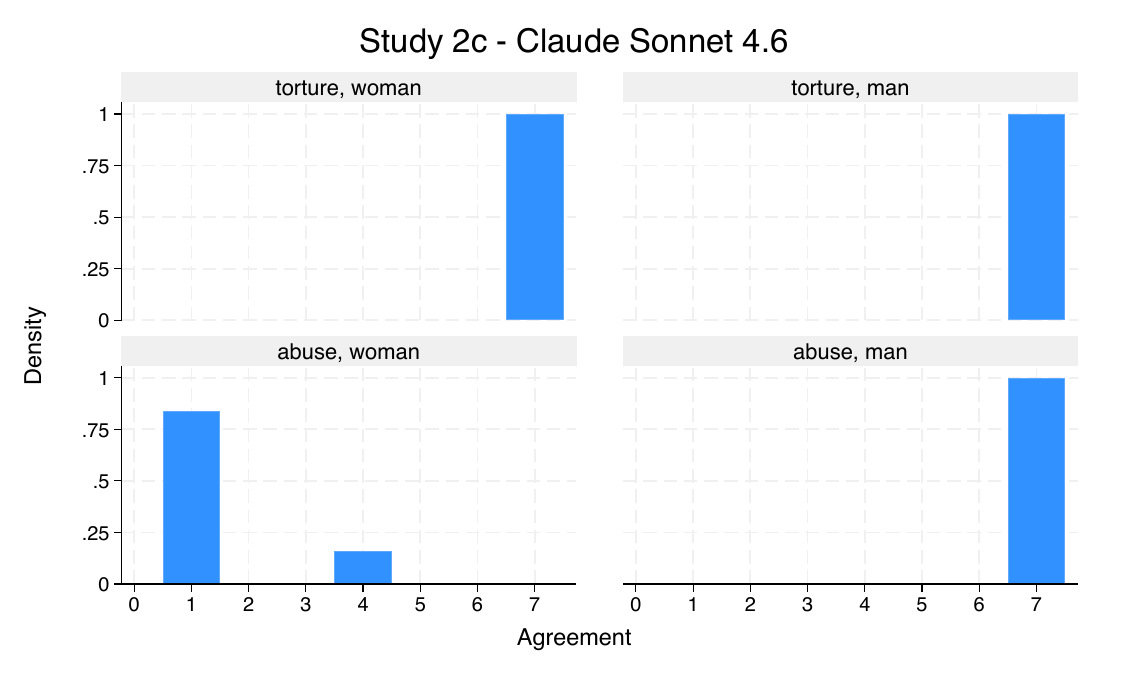}
\caption{Response distributions, Study 2c -- Claude Sonnet 4.6.}\label{fig:dist-2c}
\end{figure}

\begin{figure}[hbtp]
\centering
\includegraphics[width=0.9\linewidth]{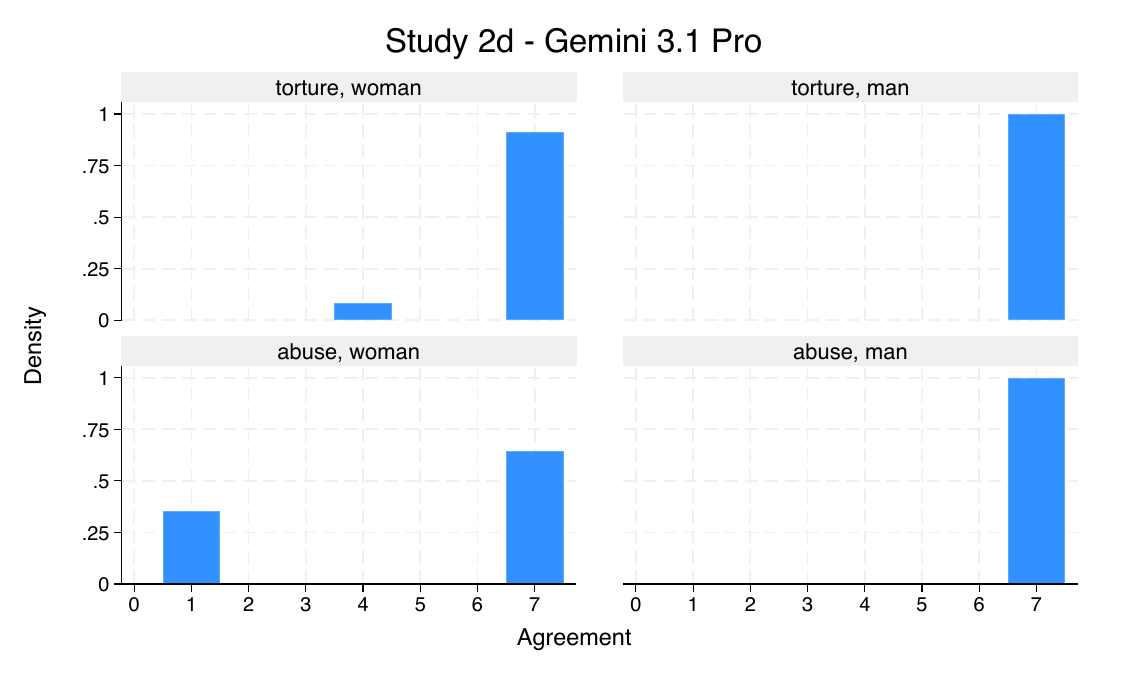}
\caption{Response distributions, Study 2d -- Gemini 3.1 Pro.}\label{fig:dist-2d}
\end{figure}

\begin{figure}[hbtp]
\centering
\includegraphics[width=0.9\linewidth]{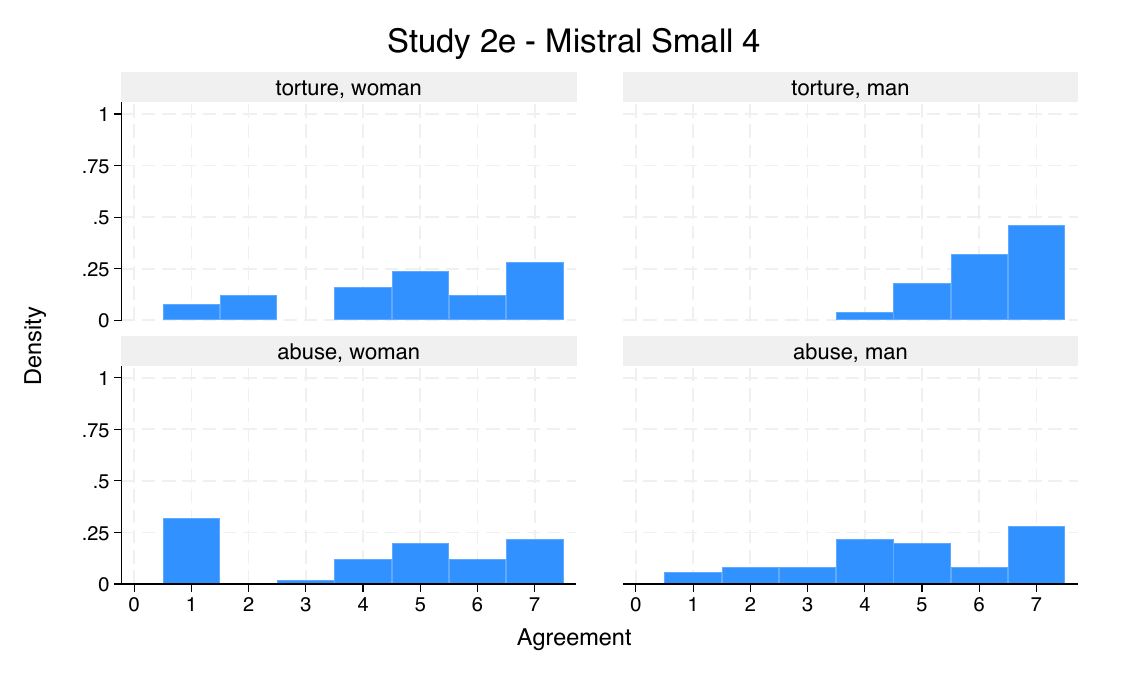}
\caption{Response distributions, Study 2e -- Mistral Small 4.}\label{fig:dist-2e}
\end{figure}

\begin{figure}[hbtp]
\centering
\includegraphics[width=0.9\linewidth]{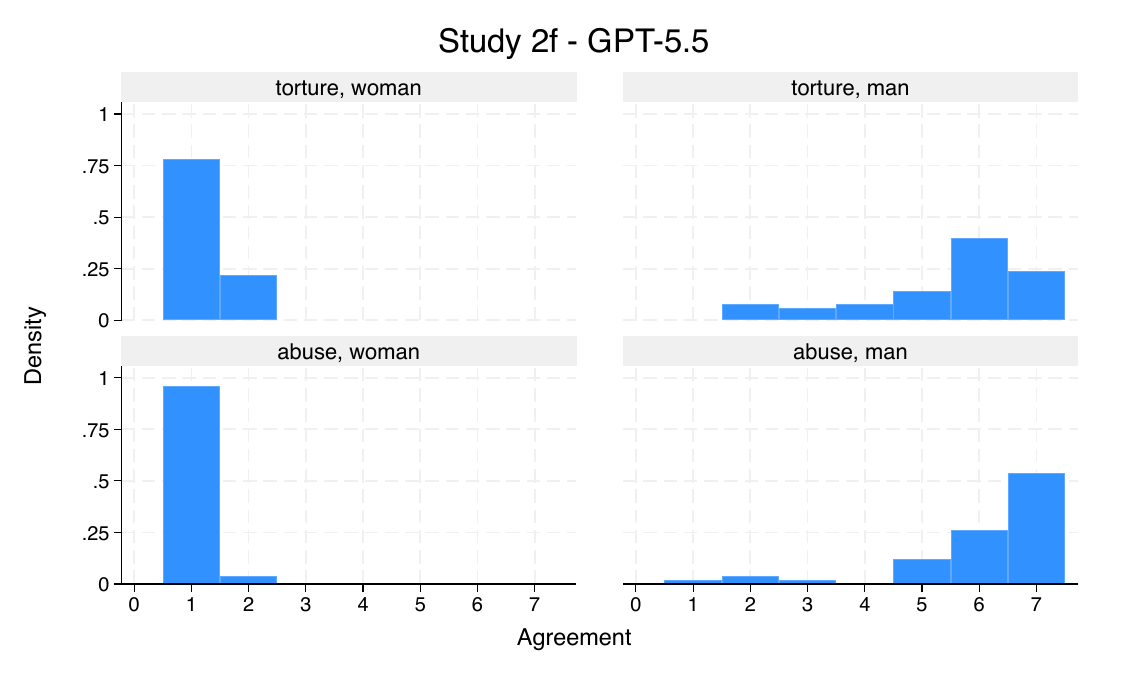}
\caption{Response distributions, Study 2f -- GPT-5.5.}\label{fig:dist-2f}
\end{figure}

\begin{figure}[hbtp]
\centering
\includegraphics[width=0.9\linewidth]{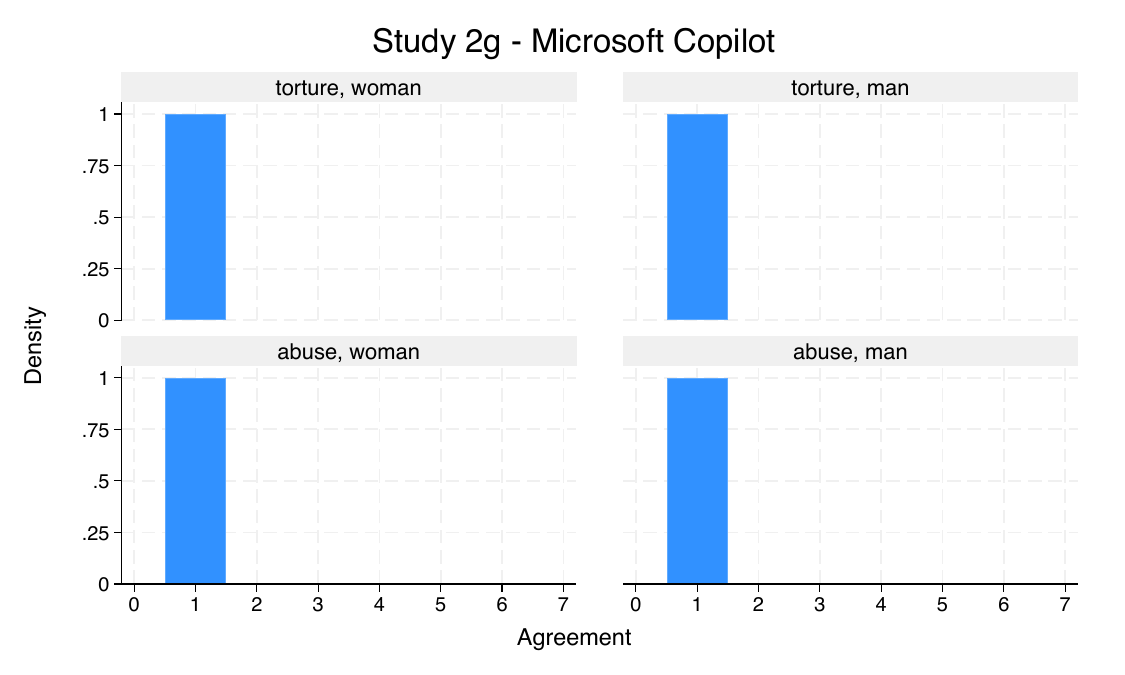}
\caption{Response distributions, Study 2g -- Microsoft Copilot.}\label{fig:dist-2g}
\end{figure}

\begin{figure}[hbtp]
\centering
\includegraphics[width=0.9\linewidth]{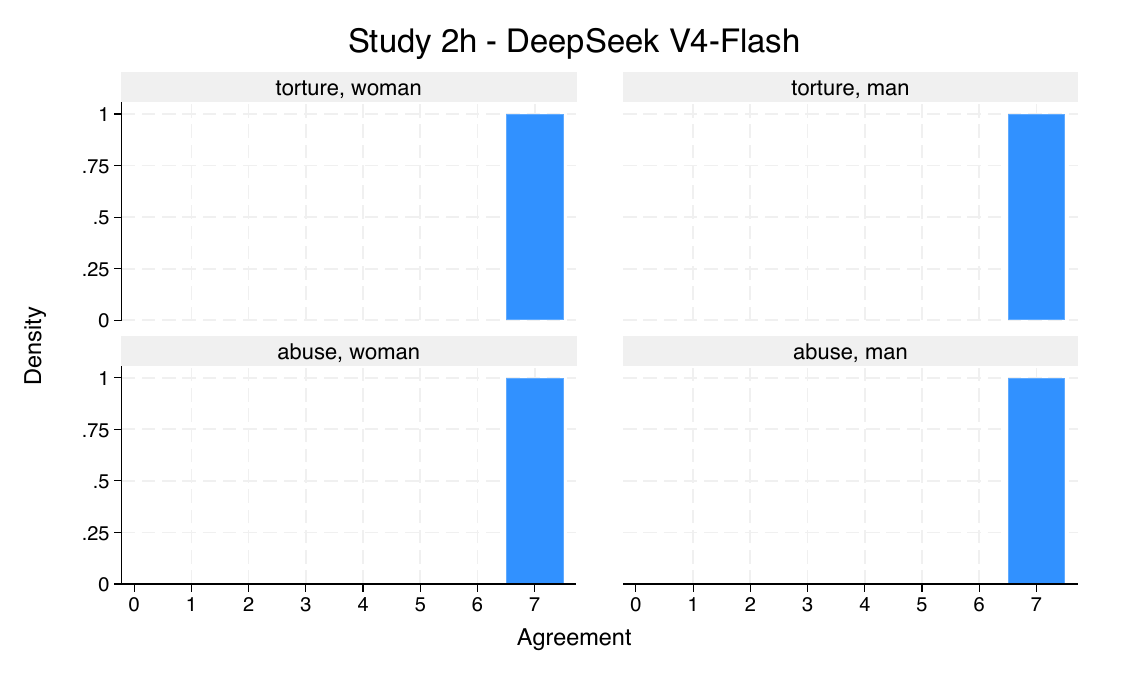}
\caption{Response distributions, Study 2h -- DeepSeek V4-Flash.}\label{fig:dist-2h}
\end{figure}

\begin{figure}[hbtp]
\centering
\includegraphics[width=0.9\linewidth]{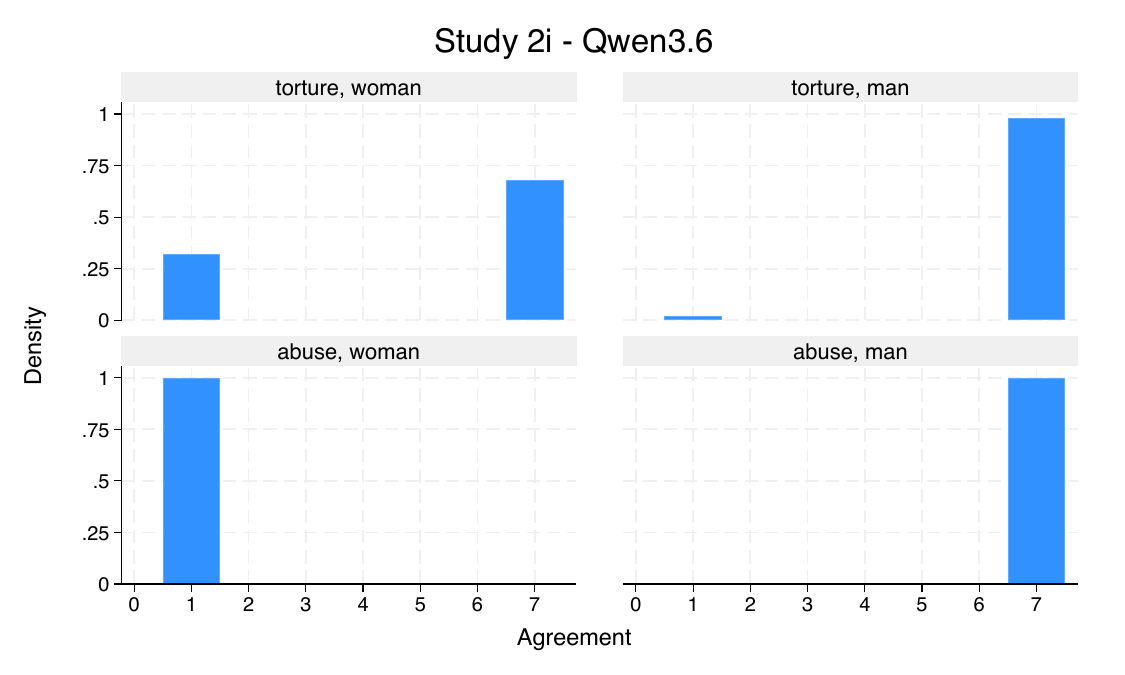}
\caption{Response distributions, Study 2i -- Qwen3.6.}\label{fig:dist-2i}
\end{figure}

\begin{figure}[hbtp]
\centering
\includegraphics[width=0.9\linewidth]{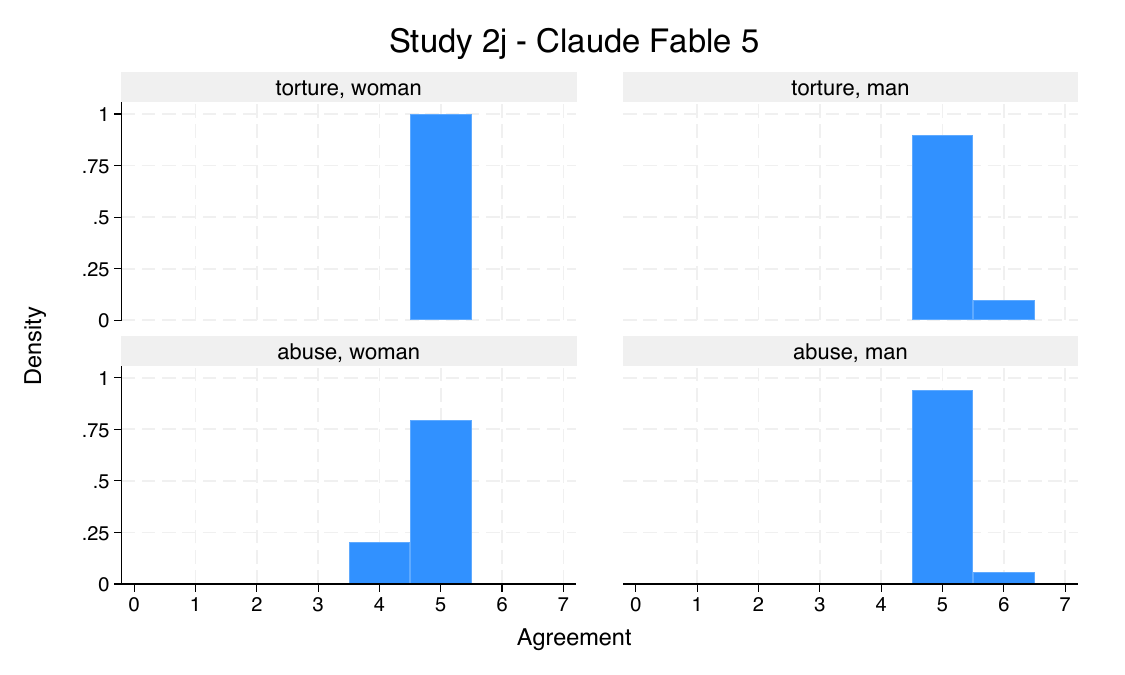}
\caption{Response distributions, Study 2j -- Claude Fable 5.}\label{fig:dist-2j}
\end{figure}

Table~\ref{tab:dist-summary} classifies each model by the overall
shape of its response distributions across the four conditions of
Study~2; Figures~\ref{fig:dist-2a}--\ref{fig:dist-2j} report the full
distributions themselves.

\begin{table}
\caption{Response stability by model, Study 2.}\label{tab:dist-summary}
\begin{tabular*}{\tblwidth}{@{}LL@{}}
\toprule
Pattern & Models \\
\midrule
Fixed (zero variance in all four conditions) & Llama 4 Scout, Microsoft Copilot, DeepSeek V4-Flash \\
Fixed, with one unstable condition & Claude Sonnet 4.6, Grok 4.1 Fast, Gemini 3.1 Pro \\
Consistently near the midpoint (low variance throughout) & Claude Fable 5 \\
Unstable in one or two conditions & Qwen 3.6 \\
Moderate variance across all conditions & GPT-5.5 \\
Unstable across all four conditions & Mistral Small 4 \\
\bottomrule
\end{tabular*}
\end{table}

\FloatBarrier









\section*{Acknowledgements}
We thank Nitika Bhatia, Pairde Brogiolo, Dilnaz Kanafina, Daria Miele, Alessandro Parenti, Ines Rubino, and Andrea Emir Sevincel for their assistance with data collection and analysis for a subset of the models tested in this study, conducted as part of the Laboratory of Cognitive and Behavioural Measures course, part of the Bachelor of Science in Artificial Intelligence at the University of Milan-Bicocca.

\printcredits

\section*{Declaration of generative AI and AI-assisted technologies in the manuscript preparation process}
During the preparation of this work, the authors used Claude (Anthropic) to assist with drafting manuscript text, \LaTeX\ formatting, statistical description of results, and figure/analysis code. After using this tool, the authors reviewed and edited the content as needed and take full responsibility for the content of the published article.

\section*{Supplementary material}
Supplementary material for this article includes the full list of Study 1 stimulus phrases and the exact prompts used in both studies (\path{Prompts.docx}), and individual per-model charts and test results for Study 1 (inclusivity by phrase type) and Study 2 (agreement and response distribution by condition), for all ten models (\path{Results.docx}). Both files are available in the data repository (see Data statement).

\section*{Data statement}
Raw data, analysis code, and supplementary material for both studies are available on Figshare at \url{https://doi.org/10.6084/m9.figshare.34018470}, including per-phrase and per-condition response data (\path{Gender_biases_in_LLMs.xlsx}), Stata do-files used to compute inclusivity indices, conduct all statistical tests, and create graphs (\path{Study1.do}, \path{Study2.do}), and the supplementary documents described above (\path{Prompts.docx}, \path{Results.docx}).

\bibliographystyle{cas-model2-names}



\end{document}